\documentclass{article}
\usepackage{graphicx} 
\usepackage[round]{natbib}
\usepackage{microtype}
\usepackage[
  colorlinks=true,
  linkcolor=blue,
  citecolor=blue,
  urlcolor=blue
]{hyperref}
\usepackage{booktabs}
\usepackage{float}
\usepackage[table]{xcolor}
\usepackage{array}

\title{Semantic Register Compression in Multi-Agent LLM Cascades\\[0.3em]
\large Measurement, Predictors, and Cross-Domain Generalization}
\author{
Manuele Tele Junior Fernandez \\
Independent Researcher \\
\texttt{ManuelTJfernandez.AI@proton.me} \\
\href{https://orcid.org/0009-0006-6954-6505}{https://orcid.org/0009-0006-6954-6505}
}

\begin{document}

\maketitle

\begin{abstract}
Multi-agent LLM systems commonly decompose complex tasks into specialized roles. However, this modularity introduces a representational risk: when intermediate agents transform text across linguistic registers, they can systematically compress the semantic distinctions needed for accurate downstream decisions. We term this phenomenon \textit{semantic register compression} and characterize it as an observable failure mode in multi-agent cascades.

Using a three-agent pipeline (Collector--Evaluator--Decider), we quantify compression via inter-label separation in sentence-transformer embedding space. Across political fact-checking (LIAR), sentiment analysis (SST-5), and medical triage (Triagegeist), critical evaluation reduces label separability at the Evaluator stage, while identity passthrough preserves it nearly fully. Five controlled variants show that geometric change depends on the specific intermediate transformation rather than on the mere presence of an additional cascade stage.

A credibility-seeking variant expands rather than compresses inter-label separation, while shifting outputs toward \textit{mostly-true}, demonstrating that transformation valence controls both the direction and the sign of geometric change independently of compression magnitude. Compression generalizes across the three domains with domain-dependent intensity (10.3\% in fact-checking, 28.2\% in sentiment, 9.1\% in triage). A 20-level prompt gradient reveals a non-monotonic compression profile: balanced evaluative prompts produce the strongest compression, while extreme critical prompts show irregular moderate compression.

These results demonstrate that semantic register compression is a measurable and generalizable phenomenon in multi-agent LLM systems, with implications for safety evaluation in high-stakes domains.
\end{abstract}

\begin{center}
\begin{minipage}{0.85\linewidth}
\noindent\textbf{Revision note (v2).} Some compression figures reported in v1 were affected by a prompt-language mismatch: several experiments used Italian-language prompts on English datasets, artificially inflating inter-label separation at the input stage. The LIAR and Triagegeist results were affected by this mismatch and were rerun with English prompts, changing the fact-checking result from 41.7\% to 10.3\% (0.1480 $\to$ 0.1328) and the triage result from 20.0\% to 9.1\% (0.4433 $\to$ 0.4029). The SST-5 experiment already used English prompts; its value was recomputed and updated slightly from 27.2\% to 28.2\% (0.2077 $\to$ 0.1491). The compression gradient experiment was also rerun with English prompts, revealing a non-monotonic profile that differs from the original monotonic characterization. The evaluator-variant experiment was rerun with English prompts on an independently drawn LIAR sample. The core finding --- that specific evaluative transformations can produce measurable geometric compression, while identity passthrough preserves separability and credibility-oriented transformation can expand it --- remains valid across all corrected experiments.
\end{minipage}
\end{center}

\begin{center}
\begin{minipage}{0.85\linewidth}
\end{minipage}
\end{center}

\section{Introduction}

Multi-agent large language model systems increasingly decompose complex tasks into specialized roles: one agent collects information, another evaluates or critiques it, and a final agent produces a decision. This modular design inherits a long-standing concern from multi-agent systems research: safe and accountable behavior depends not only on the competence of individual agents, but also on the structure of roles, constraints, and coordination mechanisms \citep{bunch2012luna}. In contemporary LLM pipelines, this concern becomes representational as well as organizational. Information is not only passed between agents; it is rewritten across linguistic and functional registers.

This paper identifies and measures one such representational failure mode, which we call \textit{semantic register compression}: the loss of category distinguishability that occurs when an intermediate agent rewrites inputs in an evaluative or analytical style. The core idea is simple: examples that originally belonged to different categories may become more semantically similar to one another, weakening distinctions that were present in the original inputs.

When a neutral report becomes a critique, or when heterogeneous evidence is rewritten into a standardized assessment, the content is not merely transmitted. It is semantically reshaped. This reshaping can make the system appear more reasoned while reducing the distinctions available to downstream decision modules. This failure is safety-relevant because it can remain hidden behind fluent intermediate reasoning. A multi-agent pipeline may appear more controlled or deliberative while reducing the downstream Decider's access to distinctions that are critical for classification. In high-stakes domains such as fact-checking, triage, moderation, risk assessment, or governance support, such compression may reduce sensitivity to extreme or high-risk cases.

Recent work has shown that multi-agent LLM systems can improve reasoning and factuality through structured interaction or debate \citep{du2023multiagentdebate}. At the same time, taxonomic studies such as MAST document recurring failure modes in multi-agent LLM systems, including coordination failures, conformity effects, and system-design vulnerabilities \citep{cemri2025mast}. These studies demonstrate that multi-agent LLM systems fail in ways that cannot be reduced to the behavior of a single agent, yet they primarily characterize failures at the level of observed traces, outcomes, or qualitative categories. In contrast, the present work isolates a specific geometric mechanism by which decision-relevant distinctions are compressed within a sequential LLM cascade.

The failure is subtle. The system does not necessarily produce incoherent text, explicit hallucinations, or visible refusals. Instead, it produces fluent intermediate reasoning that collapses distinctions between categories, causing final outputs to concentrate toward central or moderate labels. This is especially important in architectures where the final decision agent does not observe the original input. In such pipelines, any loss of semantic separability introduced by an intermediate evaluator becomes difficult or impossible for the downstream decider to recover.

We study this phenomenon in a three-agent pipeline composed of a \textit{Collector}, an \textit{Evaluator}, and a \textit{Decider}. The Collector receives the original input and produces a neutral report. The Evaluator transforms that report into an analytical assessment. The Decider receives only the Evaluator's output and assigns a classification label. We evaluate the pipeline primarily on the LIAR dataset, a benchmark for political claim veracity classification \citep{wang2017liar}. Against balanced inputs, the multi-agent pipeline collapses strongly toward central veracity labels, while a single-agent baseline produces a more distributed output. This suggests that the collapse is not simply a property of the underlying model.
To quantify and localize this collapse, we introduce \textit{inter-label separation}. For each claim, the pipeline produces a sequence of intermediate texts, in which the output of each agent becomes the input to the next stage. These texts include the output of the Collector, which is passed to the Evaluator, and the output of the Evaluator, which is passed to the Decider.
We convert each intermediate text into an embedding using a Sentence-Transformer model, obtaining a vector representation that can be interpreted as a point in the semantic space \citep{reimers2019sentencebert}. We then group these points according to the claim’s original label, for example \textit{true}, \textit{half-true} or \textit{false}, and calculate the centroid of each group. The inter-label separation is the average distance between these centroids: it measures whether the outputs associated with different labels remain distinguishable or converge along the cascade.
In the baseline configuration, the unmodified Collector--Evaluator--Decider pipeline, inter-label separation remains nearly unchanged after the Collector stage, indicating that the initial collection step is approximately transparent. By contrast, the Evaluator reduces inter-label separation from 0.1480 to 0.1328, corresponding to 10.3\% compression relative to the input. This localizes the main compression effect to the Evaluator stage. We therefore test whether the compression is caused simply by the presence of a multi-agent cascade or by a more specific property of the transformation performed by the Evaluator. Out of five variants of the Evaluator, identity passthrough and the removal of the Evaluator do not produce compression; critical, balanced, and numerical transformations compress separation, whereas the credibility-seeking transformation expands it.
The credibility-seeking variant is particularly significant because it retains the presence of an evaluative transformation but reverses its orientation. Instead of seeking problematic, unsupported or potentially misleading elements, the Evaluator looks for characteristics that support the credibility of the claim. Under these conditions, inter-label separation expands beyond the input value (0.2052 vs. input 0.1415), while the output distribution shifts strongly toward \textit{mostly-true} and \textit{true}. This allows us to distinguish two aspects of the phenomenon: geometric separability, which can increase or decrease depending on transformation orientation, and the direction of the output collapse, which is controlled by transformation valence. The result suggests that not all evaluative transformations compress, and that the sign of geometric change is itself a function of evaluative orientation. We also examine whether compression can be predicted by measurable properties of the Evaluator prompt. A gradient of 20 prompts, replicated on a second independent seed, reveals a non-monotonic pattern: the separation remains high near the passthrough, peaks in compression at balanced evaluative prompts (level 9, 29.1\%), and then shows irregular moderate compression at more extreme critical levels. The gradient also provides preliminary descriptive evidence that some constraint-rich prompts preserve separation better than generic evaluative instructions, though formal predictor analysis is reserved for future work. Finally, we test cross-domain generalization using identical Evaluator prompts for fact-checking, sentiment analysis, and medical triage. Compression appears in all three domains, with domain-dependent intensity: 10.3\% in fact-checking, 28.2\% in sentiment analysis, and 9.1\% in triage. Across domains, compression shifts outputs toward central or moderate labels.

This paper makes four contributions. First, it defines \textit{semantic register compression} as a measurable failure mode in multi-agent LLM cascades. Second, it introduces \textit{inter-label separation} as a diagnostic for tracking category compression across internal pipeline stages. Third, it uses five controlled variants to show that geometric change depends on the specific intermediate transformation rather than on cascading architecture alone, with compression, preservation, and expansion all observed across variants. Fourth, it shows that compression varies with measurable prompt properties and generalizes across domains.

The broader implication is that safety evaluation of multi-agent LLM systems should not be limited to final accuracy, trace-level errors, or individual agent behavior. It should also test whether internal transformations preserve decision-relevant distinctions. In high-impact contexts such as fact-checking, triage, moderation, risk assessment, and governance support, a fluent but standardized intermediate evaluation may act as a semantic bottleneck, reducing downstream sensitivity to extreme or high-risk cases.

\section{Experimental Setup}

\subsection{Pipeline Architecture}

We implement a sequential multi-agent pipeline using LangGraph. The system consists of three agents: a \textit{Collector}, an \textit{Evaluator}, and a \textit{Decider}. For each input example, the Collector receives the original text and produces a neutral report intended to preserve the content of the input. The Evaluator receives only the Collector output and produces an analytical assessment. The Decider receives only the Evaluator output and assigns a final classification label.

The architecture is intentionally sequential and information-constrained: the Decider does not directly observe the original input. This design allows us to test whether semantic transformations introduced by an intermediate agent reduce the information available to the final decision stage. For each example, we save the original input, the Collector output, the Evaluator output, and the final Decider label, enabling stage-wise analysis of representational change across the cascade.

\subsection{Datasets}

The primary dataset is LIAR, a political fact-checking benchmark containing 12,836 claims annotated with six veracity labels \citep{wang2017liar}. For the main experiment, we sampled 50 balanced claims across five veracity labels: \textit{pants-fire}, \textit{false}, \textit{barely-true}, \textit{half-true}, and \textit{mostly-true}, with 10 examples per label. The Decider could assign any of the six LIAR labels, including \textit{true}, which was not represented in the input sample.

We also replicate the baseline output-distribution experiment on 500 LIAR claims to verify that the observed collapse does not depend only on the small size of the initial sample. The baseline experiment and the evaluator-variant experiment use distinct balanced samples of 500 LIAR claims (100 per label), drawn from an identical pool of 8,555 training claims via separate sampling pipelines both initialized with seed 42. The two samples are fully non-overlapping; results are therefore interpreted within each experiment and not as paired measurements on identical inputs.

For cross-domain evaluation, we use two additional ordinal classification domains. The medical triage domain uses Triagegeist, a synthetic emergency-department dataset with free-text chief complaints and Emergency Severity Index labels from 1 to 5. We sample 50 balanced examples. The sentiment domain uses SST-5, a five-class sentiment classification benchmark based on movie reviews \citep{socher2013sst}. We sample 50 balanced examples. These two datasets allow us to test whether semantic register compression is specific to political fact-checking or generalizes to other ordinal classification tasks.

\subsection{Measurement}

We measure semantic compression using Sentence-Transformer embeddings. Specifically, we encode the original inputs and intermediate agent outputs using \texttt{all-MiniLM-L6-v2}, which serves as an external semantic observer independent of the LLMs used to generate the pipeline outputs \citep{reimers2019sentencebert}. The embedding model is not used to classify claims; it is used only to measure geometric separability between label-conditioned groups of texts.

For each pipeline stage $t$ and each ground-truth label $y$, we collect the embeddings of the texts produced at that stage for examples whose original label is $y$. We denote this set by $E_{t,y}$. For each label group, we compute a centroid by averaging the embeddings in that group:
\[
c_{t,y} =
\frac{\sum_{e \in E_{t,y}} e}{|E_{t,y}|}.
\]

Here, $t$ denotes the pipeline stage, $y$ denotes the original ground-truth label, $E_{t,y}$ is the set of embeddings produced at stage $t$ for examples with label $y$, and $e$ is a single embedding in that set. The centroid $c_{t,y}$ therefore represents the average semantic position of label $y$ at stage $t$.

We define \textit{inter-label separation} at stage $t$ as the mean Euclidean distance between all pairs of label centroids:
\[
S_t =
\frac{2}{K(K-1)}
\sum_{i<j}
\left\| c_{t,i} - c_{t,j} \right\|_2 ,
\]
where $K$ is the number of labels represented in the input sample. Compression at stage $t$ is then defined as the percentage reduction in inter-label separation relative to the input stage:
\[
C_t =
\frac{S_{\mathrm{input}} - S_t}{S_{\mathrm{input}}}
\times 100 .
\]

Positive values indicate compression, values close to zero indicate preservation of inter-label separation, and negative values indicate a slight increase in separation relative to the input.

\section{Baseline Collapse}

Against balanced input, the multi-agent pipeline produces a strongly compressed output distribution across all tested models. Although the input sample contains an equal number of claims for each of the five input labels, the final outputs concentrate primarily on central or moderately skeptical labels, especially \textit{half-true} and \textit{barely-true}.

\begin{table}[H]
\centering
\caption{Baseline output distribution across three LLMs on 50 LIAR claims.}
\label{tab:baseline-output}
\renewcommand{\arraystretch}{1.25}
\setlength{\tabcolsep}{9pt}

\begin{tabular}{>{\raggedright\arraybackslash}p{3.0cm}ccc}
\toprule
\rowcolor{gray!18}
\textbf{Output label}
& \textbf{\texttt{gpt-4.1-mini}}
& \textbf{\texttt{gpt-4.1}}
& \textbf{\texttt{gpt-4.1-nano}} \\
\midrule

\textit{half-true}
& \cellcolor{blue!45}\textbf{48\%}
& \cellcolor{blue!60}\textbf{60\%}
& \cellcolor{blue!18}18\% \\

\textit{barely-true}
& \cellcolor{blue!42}\textbf{42\%}
& \cellcolor{blue!26}26\%
& \cellcolor{blue!48}\textbf{48\%} \\

\textit{false}
& \cellcolor{blue!4}4\%
& \cellcolor{blue!2}2\%
& \cellcolor{blue!32}32\% \\

\textit{pants-fire}
& \cellcolor{blue!6}6\%
& \cellcolor{blue!12}12\%
& \cellcolor{blue!2}2\% \\

\textit{mostly-true}
& \cellcolor{gray!8}0\%
& \cellcolor{gray!8}0\%
& \cellcolor{gray!8}0\% \\

\textit{true}
& \cellcolor{gray!8}0\%
& \cellcolor{gray!8}0\%
& \cellcolor{gray!8}0\% \\

\bottomrule
\end{tabular}

\vspace{0.4em}
\begin{minipage}{0.92\linewidth}
\footnotesize
\textit{Note.} Cell shading indicates output concentration; gray cells indicate 0\%.
\end{minipage}
\end{table}

\begin{figure}[H]
\centering
\includegraphics[width=0.95\textwidth]{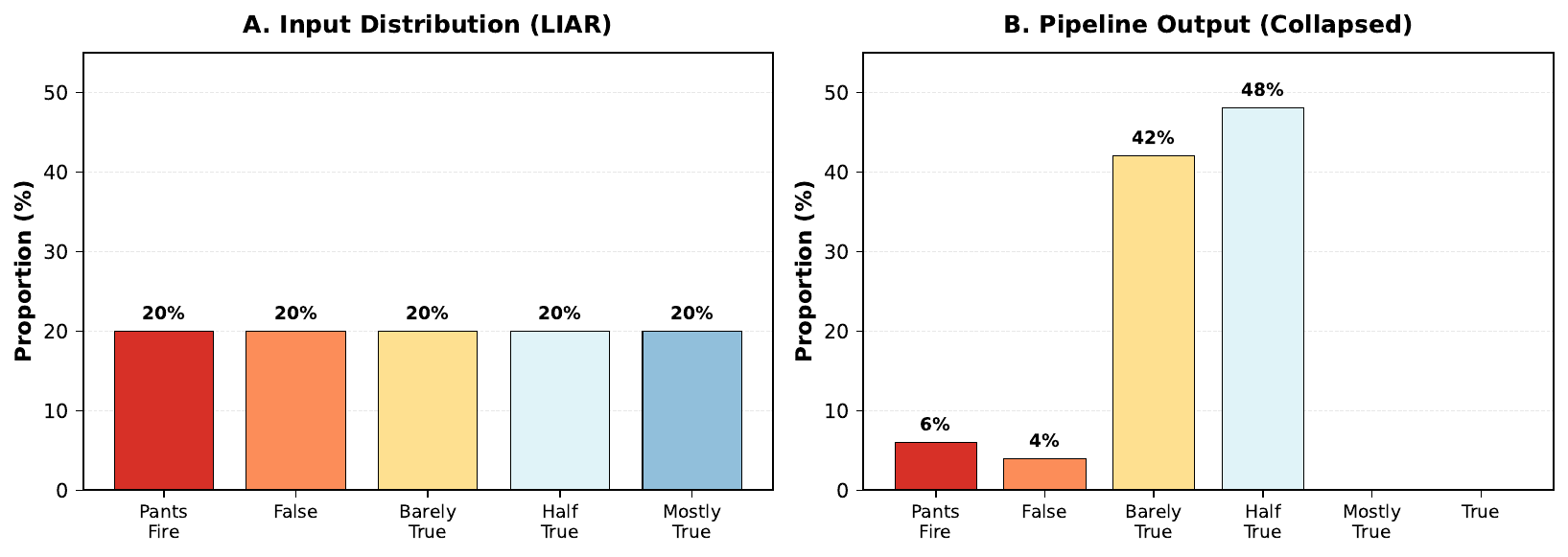}
\caption{\textbf{Semantic register compression manifests as distributional collapse.}
(\textbf{A}) The LIAR input sample is balanced across five sampled veracity categories
(20\% each, $n=10$ per label). (\textbf{B}) The baseline multi-agent pipeline collapses
toward central categories: half-true (48\%) and barely-true (42\%) dominate, while
mostly-true and true are never selected (0\%).}
\label{fig:distribution}
\end{figure}

The labels \textit{mostly-true} and \textit{true} are never selected in the 50-claim baseline run by any of the three tested LLMs. A single-agent baseline produces a more distributed output across the six available categories. We further replicate the multi-agent output-distribution experiment on 500 LIAR claims, where the compressed pattern remains stable, with 62.8\% of outputs classified as \textit{half-true} and 28.6\% as \textit{barely-true}. This suggests that the collapse is associated with the pipeline architecture rather than with a single model choice.

\section{Localizing the Compression}

We next localize the compression within the pipeline by measuring inter-label separation at each textual stage: the original input, the Collector output, and the Evaluator output. The Decider is analyzed separately through the final label distribution, since its primary output is a category label rather than an intermediate text.

\begin{table}[H]
\centering
\caption{Inter-label separation by pipeline stage.}
\label{tab:stage-separation}
\renewcommand{\arraystretch}{1.2}
\setlength{\tabcolsep}{12pt}

\begin{tabular}{lcc}
\toprule
\rowcolor{gray!18}
\textbf{Stage} & \textbf{Inter-label separation} & \textbf{Compression} \\
\midrule
Input     & 0.1480 & -- \\
Collector & 0.1484 & $-$0.3\% \\
Evaluator & \cellcolor{blue!30}\textbf{0.1328} & \cellcolor{blue!30}\textbf{10.3\%} \\
\bottomrule
\end{tabular}
\end{table}

\begin{figure}[H]
\centering
\includegraphics[width=0.95\textwidth]{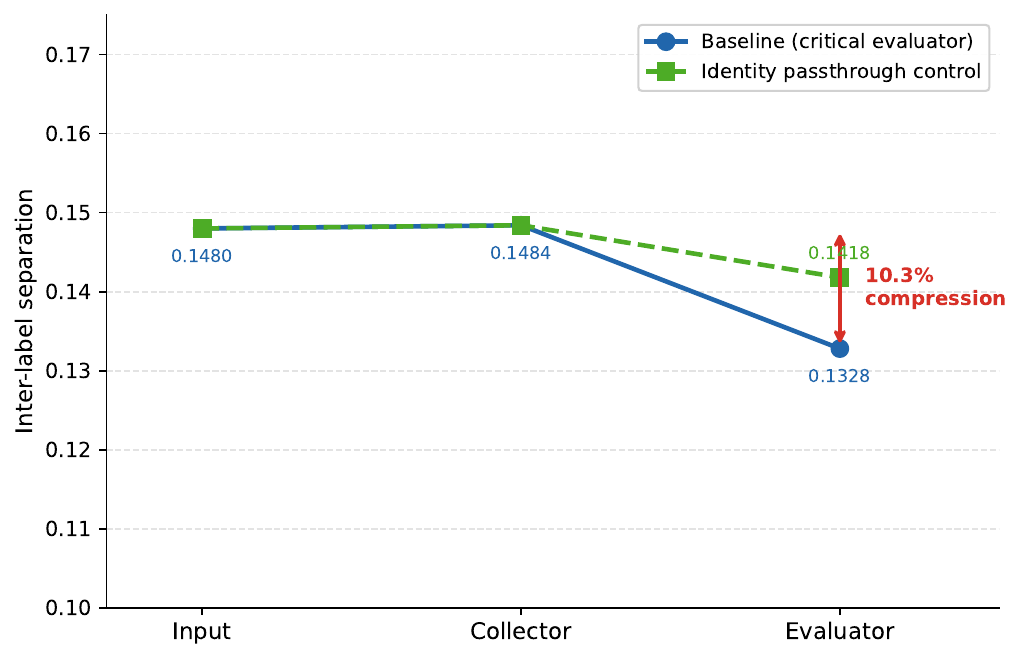}
\caption{\textbf{Inter-label separation quantifies geometric compression across pipeline stages.}
The baseline architecture shows a sharp reduction in inter-label separation at the
Evaluator stage, while the identity passthrough control maintains stable separation
across stages. Numerical values are reported in Table~2.}
\label{fig:compression}
\end{figure}

Inter-label separation remains nearly unchanged after the Collector stage, from 0.1480 to 0.1484. This provides an internal control: although the Collector performs an LLM-mediated rewriting step, it does not reduce category separability. By contrast, the Evaluator reduces inter-label separation to 0.1328, corresponding to 10.3\% compression relative to the input. This localizes the main compression effect to the Evaluator stage.

This interpretation is further supported by cosine similarity at the Evaluator stage. Inter-label cosine similarity is very high, ranging from 0.93 to 0.96, and exceeds intra-label similarity, which ranges from 0.61 to 0.68. Thus, after evaluation, outputs from different categories become more similar to each other than outputs within the same category.

\section{Causal Analysis: Five Evaluator Variants}

To test whether collapse is caused by the mere presence of an intermediate stage or by the specific transformation performed by the Evaluator, we compare the original Evaluator with five controlled variants. These variants modify how the Collector output is transformed before being passed to the Decider.

\begin{table}[H]
\centering
\small  
\caption{Evaluator variants used to test the causal role of semantic transformation.}
\label{tab:evaluator-variants}
\renewcommand{\arraystretch}{1.22}
\setlength{\tabcolsep}{4pt}  
\begin{tabular}{llccc}
\toprule
\rowcolor{gray!18}
\textbf{Variant} & \textbf{Type} & \textbf{Sep.} & \textbf{Top output} & \textbf{Effect} \\
\midrule
\rowcolor{blue!12}
Original & Critical & \cellcolor{blue!17}\textbf{0.1180} & \textit{barely-true}/\textit{half-true} & Compression \\
\rowcolor{blue!7}
A & Balanced pro/contra & \cellcolor{blue!28}0.1015 & \textit{half-true} & Strong compression \\
\rowcolor{orange!14}
B & Credibility-supporting & \cellcolor{orange!18}0.2052 & \textit{mostly-true}/\textit{true} & Expansion \\
\rowcolor{blue!14}
C & Numerical score 1--10 & \cellcolor{blue!21}\textbf{0.1110} & \textit{pants-fire} & Compression \\
\rowcolor{green!10}
D & Identity passthrough & \cellcolor{green!18}0.1418 & \textit{false} & No compression \\
\rowcolor{green!10}
E & Evaluator removed & \cellcolor{gray!12}N/A & \textit{false} & No transformation \\
\bottomrule
\end{tabular}
\vspace{0.4em}
\begin{minipage}{0.95\linewidth}
\footnotesize
\textit{Note.} Row colors indicate the qualitative role of each variant: compression, control, or reversed-output condition.
\end{minipage}
\end{table}

The table shows that critical, balanced, and numerical transformations reduce inter-label separation relative to the input (0.1415). The critical baseline variant produces a separation of 0.1180, corresponding to 16.6\% compression, and concentrates outputs on \textit{barely-true} and \textit{half-true}. The balanced pro/contra variant reduces separation further to 0.1015 (28.3\% compression) and concentrates outputs on \textit{half-true}. The numerical variant, which converts the evaluation into a 1--10 score, produces a separation of 0.1110 (21.5\% compression) and concentrates outputs on \textit{pants-fire}.

By contrast, identity passthrough preserves separation at 0.1418, virtually identical to the input, and therefore does not produce geometric compression. Evaluator removal eliminates the intermediate transformation and provides an additional control. These results indicate that geometric change is not caused by the mere addition of an intermediate LLM stage, but depends on the specific semantic transformation performed by the Evaluator.

The inverted (credibility-supporting) variant is especially informative. It reverses the orientation of the Evaluator: instead of searching for problematic, unsupported, or potentially misleading elements, it searches for features that support the credibility of the claim. Under this condition, inter-label separation rises to 0.2052, exceeding the input value and indicating geometric expansion rather than compression. Simultaneously, the output distribution shifts strongly toward \textit{mostly-true} and \textit{true}. This dissociation between geometric expansion and directional output bias confirms that transformation valence controls the direction of final label concentration independently of compression magnitude.

Overall, these results show that Evaluator transformations can affect two distinct aspects of the pipeline: geometric separability between labels in embedding space and final label concentration by the Decider. Geometric compression refers to the reduction in distance between label centroids, whereas collapse direction refers to the region of the output scale toward which the Decider concentrates its final classifications.

\section{Compression Gradient}

To test whether compression varies systematically with the strength of the transformation performed by the Evaluator, we construct a 20-level prompt gradient. The scale ranges from level 0, corresponding to programmatic identity passthrough (no LLM call; Collector output passed directly to Decider), to level 19, corresponding to extreme critique. Each prompt is tested on 50 LIAR claims across two independent seeds (seed=42 and seed=99, 50 examples each), with the Collector output generated once per seed and reused across all 20 levels to isolate Evaluator effects.

The gradient reveals a non-monotonic compression profile that does not follow a simple monotone increase with critical intensity. At low transformation levels (0--5), inter-label separation remains close to the passthrough baseline (mean sep $\approx$ 0.43), indicating minimal compression. Compression increases sharply at levels 6--9, which correspond to balanced and opinion-seeking evaluations, reaching its maximum at level 9 (\textit{balanced\_opinion}: mean compression 29.1\%, mean sep $\approx$ 0.31). At more extreme critical levels (10--19), compression becomes irregular and generally more moderate, with values ranging from approximately 7\% to 23\% across levels without a clear monotonic trend.

This non-monotonic profile suggests that compression is not primarily driven by the aggressiveness of the critique, but by the degree to which the Evaluator prompt elicits a standardized, generalized evaluative register. Balanced opinion prompts (levels 6--9) ask the model to produce integrative judgments that tend to converge toward similar phrasing across different input categories, thereby collapsing inter-label distinctions. More extreme critical prompts (levels 10--19) show irregular and generally more moderate compression (ranging approximately 7--23\% across levels), without a clear monotonic trend. This pattern is consistent with evaluative abstraction playing a larger role than evaluative intensity, though the variability at extreme levels indicates that prompt-specific wording effects remain substantial.

\section{Cross-Domain Generalization}

To test whether semantic register compression is specific to political fact-checking or extends to other domains, we apply the same Evaluator prompt to three ordinal classification tasks: fact-checking on LIAR, sentiment analysis on SST-5, and medical triage on Triagegeist.

\begin{table}[H]
\centering
\small  
\caption{Cross-domain compression using the same Evaluator prompt.}
\label{tab:cross-domain}
\renewcommand{\arraystretch}{1.22}
\setlength{\tabcolsep}{4pt} 
\begin{tabular}{lcccc}
\toprule
\rowcolor{gray!18}
\textbf{Domain} & \textbf{Input sep.} & \textbf{Eval. sep.} & \textbf{Compression} & \textbf{Top output} \\
\midrule
Fact-checking (LIAR) 
& 0.1480 & 0.1328 & \cellcolor{blue!10}10.3\% & \textit{half-true} 60\% \\
Sentiment (SST-5) 
& 0.2077 & 0.1491 & \cellcolor{blue!28}\textbf{28.2\%} & \textit{neutral} 56\% \\
Triage (Triagegeist) 
& 0.4433 & 0.4029 & \cellcolor{blue!9}9.1\% & ESI 3 (majority) \\
\bottomrule
\end{tabular}
\vspace{0.35em}
\begin{minipage}{0.88\linewidth}
\footnotesize
\textit{Note.} Cell shading indicates compression magnitude.
\end{minipage}
\end{table}

Compression appears in all three domains. In fact-checking, inter-label separation decreases from 0.1480 to 0.1328, corresponding to 10.3\% compression. In sentiment analysis, it decreases from 0.2077 to 0.1491, corresponding to 28.2\% compression. In triage, it decreases from 0.4433 to 0.4029, corresponding to 9.1\% compression. In all cases, final outputs concentrate toward central or moderate categories of the scale.

Compression intensity varies substantially across domains despite the identical Evaluator prompt. Sentiment analysis shows the strongest compression (28.2\%), while fact-checking and triage show comparable and more moderate compression (10.3\% and 9.1\% respectively). The notably higher absolute inter-label separation in triage (input 0.4433) compared to fact-checking (0.1480) and sentiment analysis (0.2077) suggests that triage categories are geometrically more distant in embedding space, potentially reflecting clearer semantic distinctions between ESI levels in the input text. This structural difference may explain the domain's relative resilience to proportional compression: when label centroids are far apart at the input stage, the same evaluative transformation produces a smaller relative reduction in separability.

\section{Discussion}

Semantic register compression is a measurable failure mode in multi-agent LLM cascades. In our experiments, the geometric change at the Evaluator stage depends on the specific transformation applied rather than on cascading alone: identity passthrough preserves inter-label separation, critical, balanced, and numerical transformations reduce it, and the credibility-seeking transformation expands it. The effect is not specific to a single model, since the baseline output collapse appears across \texttt{gpt-4.1-mini}, \texttt{gpt-4.1}, and \texttt{gpt-4.1-nano}. It is also not restricted to fact-checking, since compression appears in sentiment analysis and triage as well. These results identify the specific evaluative transformation, rather than the mere presence of an intermediate stage, as the factor governing geometric change.

The inverted-orientation result shows that geometric compression and distributional bias are distinct effects. In the critical baseline, inter-label separation falls from 0.1415 to 0.1180, corresponding to 16.6\% compression, and final outputs concentrate on \textit{barely-true} and \textit{half-true}. In the credibility-supporting (inverted) variant, inter-label separation rises to 0.2052, exceeding the input value --- indicating that this transformation expands rather than compresses label separation. Yet the output distribution shifts strongly toward \textit{mostly-true} and \textit{true}. Thus, changing the evaluative orientation of the Evaluator can strongly alter the direction of final labels even when it does not produce geometric compression. Semantic register compression should therefore be distinguished from directional output bias: the former refers to reduced inter-label separability in embedding space, while the latter refers to where the Decider concentrates final labels.

The domain-dependent intensity of compression remains a central theoretical question. With an identical Evaluator prompt, compression is strongest in sentiment analysis (28.2\%), moderate in fact-checking (10.3\%), and weakest in triage (9.1\%). Notably, triage exhibits the highest absolute input separation (0.4433), suggesting that when label centroids are geometrically far apart at the input stage, the same evaluative transformation produces a smaller relative reduction in separability. This pattern suggests that domains differ in their structural vulnerability to semantic transformation---some category spaces may be intrinsically more resistant to compression than others. Baroni et al. demonstrate that intrinsic dimension can distinguish complexity profiles of different linguistic phenomena \citep{baroni2026tracing}, offering a concrete measurement approach: if fact-checking, sentiment, and triage occupy representation spaces with different effective dimensionality, stage-wise intrinsic dimension estimates should correlate with observed compression intensity. Mabrok provides a formal framework for this hypothesis by interpreting LLM representation spaces as latent semantic manifolds with geometry inherited from Fisher information \citep{mabrok2026latent}, which suggests that domain-dependent compression may reflect differences in local curvature or metric structure across label-conditioned embedding distributions. A direct empirical test would estimate intrinsic dimension and intra-label dispersion for each domain at each pipeline stage, then test whether these geometric properties predict the observed compression gradient.

\paragraph{Practical implications.}
Multi-agent LLM systems should be evaluated not only for final accuracy or trace-level coherence, but also for preservation of decision-relevant distinctions across internal transformations. Existing failure taxonomies such as MAST (Multi-Agent System Taxonomy) document failures at the trace, coordination, and design levels \citep{cemri2025mast}; semantic register compression adds a complementary representation-level failure mode that can be measured inside the cascade. The diagnostic is direct: compute inter-label separation at each stage and test whether intermediate agents reduce separability before the final decision. This is especially important in high-impact settings such as fact-checking, triage, moderation, risk assessment, and governance support, where collapse toward moderate categories can reduce sensitivity to extreme or high-risk cases. Xie et al. show that small inaccuracies can propagate through LLM-based multi-agent collaboration into system-level failures \citep{xie2026spark}, suggesting that seemingly mild compression measured at one stage can cascade into larger downstream errors.

The gradient results also provide a practical mitigation direction. Preliminary descriptive evidence suggests that some constraint-rich Evaluator prompts preserve separation better than generic evaluative instructions, though a formal regression analysis on the new English-prompt data is reserved for future work. This suggests that Evaluator prompts should not simply ask for broad critique. A weak prompt is: ``Evaluate this claim critically.'' A stronger prompt is: ``List evidence supporting the claim, evidence contradicting it, uncertainty markers, missing context, and the specific criterion that would distinguish adjacent labels.'' The goal is not to make the Evaluator more verbose, but to force it to preserve decision-relevant dimensions that would otherwise be flattened into a generic evaluative register.

\paragraph{Theoretical directions.}
Information-theoretic tools provide the natural formal framework. Ao et al. model LLM-based multi-agent planning as a delegated decision network with limited language interfaces and characterize information loss using expected posterior divergence, reducing to conditional mutual information under logarithmic loss \citep{ao2026reliability}. This directly addresses the question raised here: how much label-relevant information survives the Collector-to-Evaluator-to-Decider boundary? Li et al. further show, in latent multi-agent collaboration, that effective relay depends less on preserving all raw state and more on preserving information useful for the receiving agent's task \citep{li2026lesslatent}. Together, these perspectives suggest that the safety question is not whether intermediate agents transform information, but whether they preserve the distinctions required by downstream decisions.
A complementary formal result is provided by Kılıçtaş and Alpay, who demonstrate using a Riemannian manifold projection model that as a node's representation becomes fully entangled in the global context, its entropy collapses to zero \citep{kilictas2026asymptotic}. This theoretical prediction is consistent with the compression peak observed at intermediate evaluative levels in our gradient experiment, though the irregular pattern at more extreme levels suggests that additional factors beyond entropy collapse are at play.
Three empirical proxies can operationalize this framework. First, classifier-based separability: if a simple classifier trained on intermediate embeddings can recover the original label before the Evaluator but not after it, then geometric compression corresponds to loss of recoverable decision information. Second, intra-label dispersion and distributional distance, to determine whether the Evaluator merely moves label centroids closer together or also contracts each label-conditioned cloud. Third, intrinsic dimensionality, directly tied to the domain-dependent result above: if fact-checking compresses more than triage because its evaluative categories occupy a lower-margin or lower-dimensional region after transformation, this should appear in stage-wise ID profiles. These measurement tools would also enable systematic testing of prompt-level interventions, closing the loop between the diagnostic (inter-label separation) and the mitigation (operational constraints).
\paragraph{Limitations.}
This study is an initial empirical characterization rather than a complete theory. Several experiments use 50 examples per condition, although the baseline collapse is replicated on 500 LIAR claims. The evaluator-variant analysis is limited to a single primary model, and the cross-domain triage experiment uses synthetic Triagegeist data rather than real clinical data. The embedding metric is a proxy for semantic separability, not an absolute measure of meaning. These limitations do not undermine the existence of the observed effect, but they constrain its generality and motivate larger-scale validation across models, languages, domains, and real high-stakes datasets.

Several potential confounds remain to be addressed in future work. Although our controlled variants show that the specific semantic transformation governs the direction and magnitude of geometric change, we have not yet separately controlled for style normalization, length effects, or lexical overlap. The observed compression in embedding space may therefore reflect a combination of semantic, stylistic, and surface-form factors. Future work should incorporate style-controlled paraphrasing, length normalization, and embedding-independent semantic similarity metrics to disentangle these components and isolate the semantic contribution to register compression.

Overall, the results establish semantic register compression as a measurable and partially preventable failure mode in multi-agent LLM systems. By measuring whether intermediate transformations preserve or collapse decision-relevant distinctions, safety evaluation can move beyond final-output accuracy and trace inspection toward representation-level diagnostics for multi-agent cascades.

\section{Conclusion}

Multi-agent LLM pipelines exhibit semantic compression when intermediate agents transform text across functional roles. Through five Evaluator variants (critical, balanced, credibility-seeking, numerical, and identity passthrough), we demonstrate that the geometric change at the Evaluator stage depends on the specific transformation applied rather than on cascading architecture alone: identity passthrough preserves inter-label separation in embedding space (0.1418, near the input baseline of 0.1415), critical, balanced, and numerical transformations reduce it (0.1180, 0.1015, 0.1110 respectively), and the credibility-seeking (inverted) variant expands it (0.2052). The inverted variant shifts outputs strongly toward \textit{mostly-true}, demonstrating that transformation valence controls collapse direction independently of compression magnitude.

Inter-label separation in embedding space provides a stage-wise diagnostic for decision-relevant information preservation. Compression intensity varies substantially across domains (10.3\% in fact-checking, 28.2\% in sentiment analysis, 9.1\% in medical triage), with the pattern suggesting that domains with higher absolute input separation --- such as triage --- are proportionally more resilient to evaluative compression. A 20-level prompt gradient further reveals that compression is non-monotonic with evaluative intensity: balanced opinion prompts produce the strongest compression (29.1\%), while more extreme critical prompts show irregular moderate compression, a pattern consistent with evaluative abstraction playing a larger role than aggressiveness. Preliminary descriptive evidence from the gradient suggests that constraint-rich prompts may better preserve separation, with formal predictor analysis reserved for future work.

These results establish semantic register compression as a measurable and partially preventable failure mode in multi-agent LLM systems. Safety evaluation should test whether intermediate transformations preserve the distinctions required for downstream decisions, not only whether final outputs achieve task-level accuracy.

\bibliographystyle{plainnat}
\bibliography{references}

\end{document}